\documentclass{article}

\usepackage{PRIMEarxiv}

\usepackage[utf8]{inputenc} %
\usepackage[T1]{fontenc}    %
\usepackage{hyperref}       %
\usepackage{amsfonts}       %
\usepackage{nicefrac}       %
\usepackage{microtype}      %
\usepackage{lipsum}
\usepackage{fancyhdr}       %
\usepackage{graphicx}       %
\graphicspath{{media/}}     %

\usepackage{color}
\usepackage{booktabs}
\usepackage{multirow}
\usepackage{mathtools}
\usepackage{url}

\newcommand{\figwidth}[4]{
    \begin{figure}[tb]\centering
    \includegraphics[width=#4]{./fig/#1}
    \caption{#2}
    \label{#3}\end{figure}
    {}}

\newcommand{\figccwidth}[4]{
    \begin{figure*}[tb]\centering
    \includegraphics[width=#4]{./fig/#1}
    \caption{#2}
    \label{#3}\end{figure*}
    {}}

\title{From Volumes to Slices: Computationally Efficient Contrastive Learning for Sequential Abdominal CT Analysis
}

\author{
  Po-Kai Chiu, Hung-Hsuan Chen \\
  Computer Science \& Information Engineering\\
    National Central University\\
	Taoyuan, Taiwan
  \texttt{tyjh22005@gmail.com, hhchen1105@acm.org} \\
}

\begin{document}
\maketitle

\begin{abstract}

The requirement for expert annotations limits the effectiveness of deep learning for medical image analysis. Although 3D self-supervised methods like volume contrast learning (VoCo) are powerful and partially address the labeling scarcity issue, their high computational cost and memory consumption are barriers. We propose 2D-VoCo, an efficient adaptation of the VoCo framework for slice-level self-supervised pre-training that learns spatial-semantic features from unlabeled 2D CT slices via contrastive learning. The pre-trained CNN backbone is then integrated into a CNN-LSTM architecture to classify multi-organ injuries. In the RSNA 2023 Abdominal Trauma dataset, 2D-VoCo pre-training significantly improves mAP, precision, recall, and RSNA score over training from scratch. Our framework provides a practical method to reduce the dependency on labeled data and enhance model performance in clinical CT analysis. We release the code for reproducibility.\footnote{\url{https://github.com/tkz05/2D-VoCo-CT-Classifier}}

\end{abstract}


\section{Introduction}
\label{sec:intro}

Traumatic abdominal injury is a critical medical condition in which a rapid and accurate diagnosis is essential. Although computed tomography (CT) is the primary imaging modality for this task, the manual review process is time-consuming and requires expert radiologists, who are often in short supply. Deep learning has shown immense promise in automating medical image analysis, offering a pathway to faster, more consistent diagnostic support. However, its success depends heavily on the need for large-scale, meticulously annotated datasets, which is a significant bottleneck due to the high cost of labeling by specialized experts. To mitigate this dependency, self-supervised learning (SSL) has emerged as a powerful paradigm, enabling models to learn rich feature representations from unlabeled data to boost performance in downstream clinical tasks.

Despite the potential of SSL, there is still a primary challenge in applying these methods effectively to volumetric medical data. Abdominal CT analysis is complex due to intricate anatomy, high variability between patients, and subtle visual cues of traumatic injury. State-of-the-art SSL frameworks designed for volumetric data, such as 3D Volume Contrastive Learning (VoCo)~\cite{wu2024voco}, effectively capture this spatial context but incur prohibitive computational costs. Their reliance on 3D convolutions demands significant memory and processing power, creating a barrier to widespread adoption in typical clinical environments. This creates a clear research gap: the need for a method that retains the semantic learning power of 3D SSL while operating within the computational constraints of a 2D framework.

To bridge this gap, this study aims to develop and validate a computationally efficient SSL framework that adapts the core principles of 3D VoCo for 2D slice-level pre-training. This paper introduces 2D-VoCo, an adaptation of the VoCo framework designed for sequential 2D medical images. Our approach significantly improves classification performance in the RSNA 2023 abdominal trauma dataset~\cite{ma2024automatic} compared to standard supervised training. Ultimately, this work presents a practical solution that merges the power of spatial representation learning with the efficiency of 2D workflows, providing an effective tool to enhance clinical AI models while reducing the reliance on scarce annotated data.

\section{Method} \label{sec:method}

\figccwidth{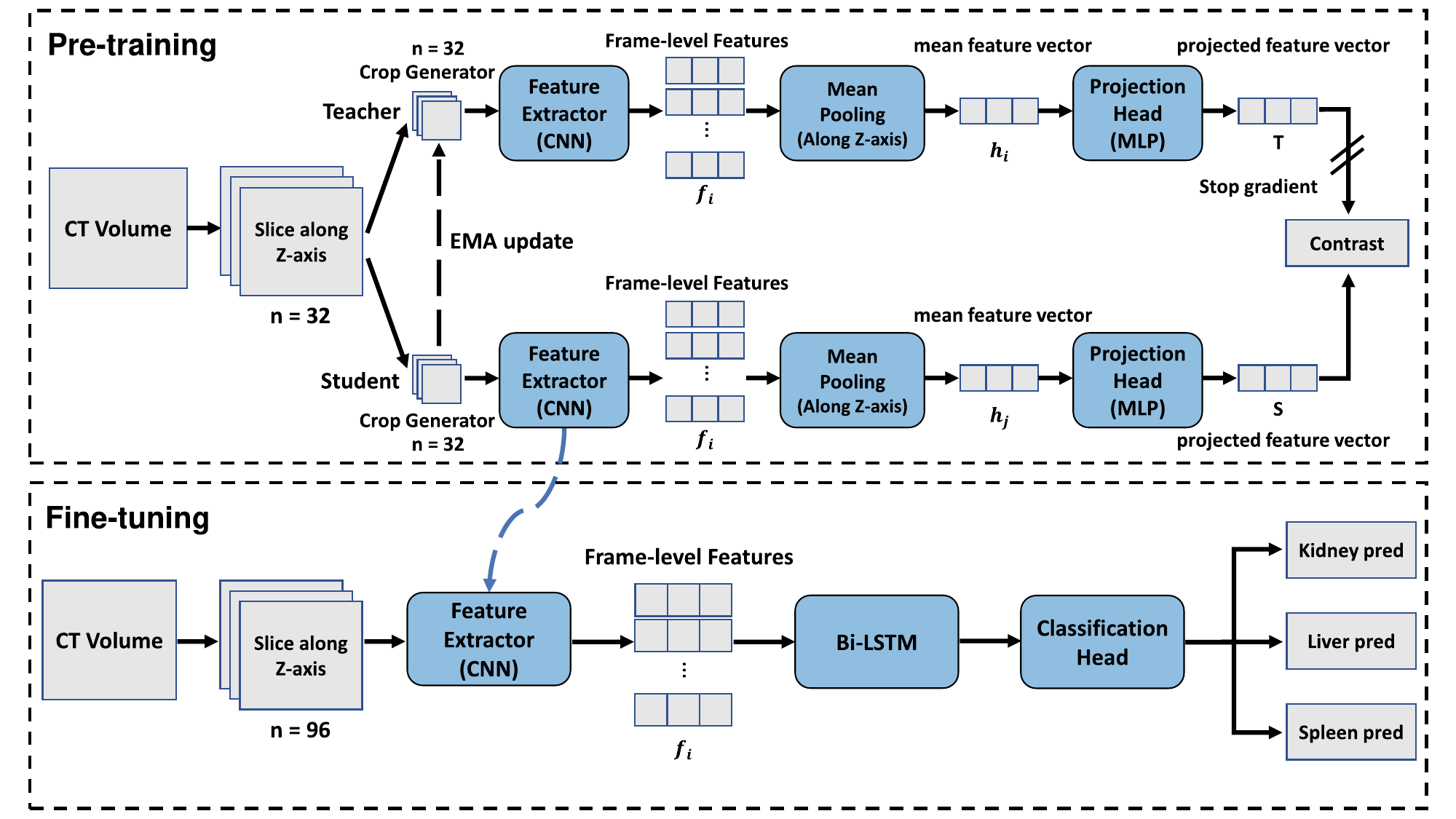}{Overall Framework: the process includes an upstream pre-training stage with the 2D-VoCo student-teacher framework and a downstream fine-tuning stage using a CNN-LSTM architecture for classification.}{fig:overall-framework}{\textwidth}

\figwidth{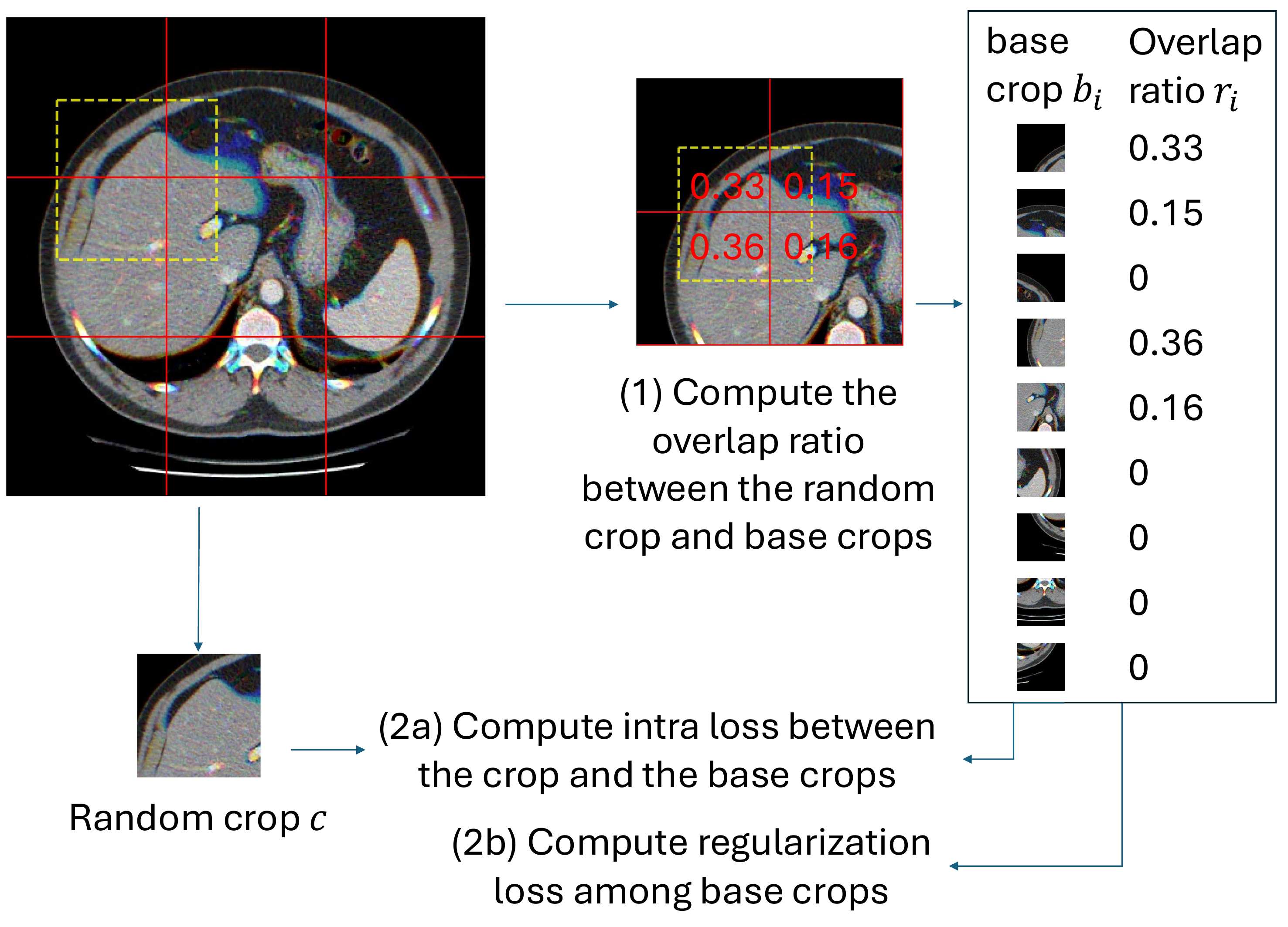}{Intra loss and regularization loss for 2D VoCo pre-training.}{fig:2dvoco-pretrain-intra-loss}{.6\columnwidth}

Our proposed method follows a two-stage learning process, as illustrated in Figure~\ref{fig:overall-framework}. The first stage involves self-supervised pre-training using our 2D-VoCo framework on unlabeled abdominal CT scans to learn robust slice-level feature representations. The pre-trained feature extractor is transferred and fine-tuned in the second stage on a downstream task for multi-organ abdominal trauma classification using a sequence-aware architecture.

\subsection{Upstream Pre-training: The 2D-VoCo Framework}

To address the computational limitations of 3D models, we adapt the VoCo framework for 2D slice-based analysis. Our 2D-VoCo approach preserves the core concept of learning spatial-semantic context through contrastive learning, but operates efficiently on sequences of 2D CT slices. 

The framework employs a momentum-based student-teacher architecture. Both branches share an identical 2D CNN backbone (EfficientNetV2)~\cite{tan2021efficientnetv2}. The student model's weights are updated via backpropagation, while the teacher model's weights are an exponential moving average (EMA) of the student's weights. The teacher branch does not receive gradient updates, which helps stabilize the learning process.

For self-supervised pre-training, we generate samples by creating a grid of non-overlapping base crops ($b_i$) and a randomly positioned random crop ($c$) from the axial slices of each 3D CT volume (referring to Figure~\ref{fig:2dvoco-pretrain-intra-loss}). These crops are processed through a student-teacher CNN architecture to produce the random crop embeddings $f_c$ and the base crop embedding $f_{b_i}$. The training objective is to make the cosine similarity between these embeddings, $\cos(f_c, f_{b_i})$, approximate their spatial overlap $r_i$, defined by the Intersection over Union (IoU) between the crops. The model learns this relationship by minimizing the following intra loss.

\begin{equation} \label{eq:slice-loss}
\mathcal{L}_{\text{intra}}(f_c, f_{b_i}) \coloneq -\log\left(1 - \left|r_i - \cos(f_c, f_{b_i})\right|\right).
\end{equation}

To improve feature consistency between patients, we align the cosine similarities between a base crop and a random crop between patients  (e.g., patients $A$ and $B$).

\begin{equation}
     \mathcal{L}_\text{inter}(f_c, f_{b_i}, f'_c, f'_{b_i}) \coloneq 
     -\log\left(1-\left|\cos(f_c, f_{b_i}) - \cos(f'_c, f'_{b_i})\right|\right),
\end{equation}
where the random crop $c$ is from a patient $A$ and the base crops $b_i$ are from a patient $B$, $f_c$ and $f_{b_i}$ are the embeddings of a random crop $c$ and a base crop $b_i$ from the teacher model, and $f'_c$ and $f'_{b_i}$ are from the student model.

To prevent the features of base crops from collapsing, we encourage distinct embeddings for each base crop.

\begin{equation}
    \mathcal{L}_\text{reg}(f_{b_i}, f_{b_j}) \coloneq |\text{cos}(f_{b_i}, f_{b_j})|
\end{equation}

The total loss is the sum of $\mathcal{L}_\text{intra}$, $\mathcal{L}_\text{inter}$, and $\mathcal{L}_\text{reg}$.

\subsection{Downstream Fine-tuning: Classification}

For the downstream task, the student model's pre-trained 2D CNN backbone is used as the feature extractor.
Each CT scan is processed by the pre-trained CNN to extract a sequence of feature vectors. This sequence is then fed into a bidirectional long short-term memory (Bi-LSTM) to model the spatial dependencies and contextual relationships between adjacent slices. The Bi-LSTM output is passed to a multi-head classification layer, which has separate heads to predict the injury status (healthy, low injury, or high injury) for the kidney, liver, and spleen.

\section{Experiments}

\subsection{Datasets and Pre-processing}

Our primary dataset for the downstream classification task was the RSNA 2023 Abdominal Traumatic Injury CT (RATIC) dataset~\cite{rudie2024rsna}. It contains $3,147$ patient CT scans with labels for kidney, liver, and spleen injuries, each categorized as healthy, low injury, or high injury. The dataset was split by patient ID into an $80\%$ training set and a $20\%$ test set, and a 5-fold cross-validation was performed on the training set to determine the threshold. 

Data preparation was tailored for each stage. For upstream pre-training, CT volumes were resampled to a uniform size (32, 192, 192), i.e., 32 slices, each of size $192 \times 192$. Non-abdominal regions were removed to focus on relevant anatomy. For the downstream classification task, the DICOM files were processed with appropriate windowing to enhance organ contrast, resampled to a fixed size of (96, 256, 256).

\subsection{Experimental Setup}

To first validate the practicality of the original 3D VoCo, we attempted to reproduce its 3D pre-training process based on the original paper and the official code. However, it was computationally prohibitive; our single NVIDIA RTX 3090 (24GB) GPU could not support the recommended batch size of 4. A minimal batch size of 1 was impractically slow (approx. 1 hour/epoch vs. 20 minutes/epoch for our 2D model at the same batch size). This confirmed the need for a more efficient 2D adaptation, motivating our subsequent experiments.

We used two 2D CNN backbones: EfficientNetV2-T and EfficientNetV2-S. The experimental models were pre-trained using our 2D-VoCo framework and then fully fine-tuned on the downstream task. The baseline model (without 2D VoCo) was trained from the ImageNet pre-trained weights. 

Model performance was assessed using an official competition metric, the RSNA score (a weighted log loss), along with standard classification metrics: mean Average Precision (mAP), Precision, and Recall. Precision and recall were macro-averaged across all classes and the three organs.

\subsection{Results and Analysis}

\begin{table*}[tbh]
\centering
\caption{Classification result comparison for baseline, 2D VoCo (without extra images from FLARE23), and 2D VoCo (with unlabeled FLARE23 images for pre-training). Best and second best are highlighted in bold and underlined, respectively.}
\label{tab:voco_experiment2}
\begin{tabular}{cc|cccc}
\toprule
\textbf{Model} & \textbf{Setting} 
& \textbf{RSNA Score $\downarrow$} & \textbf{mAP (\%) $\uparrow$}
& \textbf{Precision (\%) $\uparrow$} & \textbf{Recall (\%) $\uparrow$} \\
\midrule
\multirow{3}{*}{EffNetV2-T}
& w/o 2D VoCo        & 0.4133 $\pm$ 0.0068 & \underline{59.47 $\pm$ 0.01} & 60.84 $\pm$ 0.55 & 49.37 $\pm$ 0.25 \\
& 2D VoCo & \underline{0.3818 $\pm$ 0.0052} & 58.71 $\pm$ 0.83 & \underline{65.88 $\pm$ 1.02} & \underline{53.13 $\pm$ 0.60} \\
& 2D VoCo w/ extra unlabled data & \pmb{0.3777 $\pm$ 0.0084} & \pmb{60.98 $\pm$ 1.09} & \pmb{66.71 $\pm$ 0.19} & \pmb{53.90 $\pm$ 0.15} \\
\midrule
\multirow{3}{*}{EffNetV2-S}
& w/o 2D VoCo        & 0.4009 $\pm$ 0.0054 & 59.20 $\pm$ 0.44 & 62.28 $\pm$ 0.73 & \underline{52.05 $\pm$ 2.00} \\
& 2D VoCo & \underline{0.3925 $\pm$ 0.0068} & \underline{60.13 $\pm$ 0.60} & \underline{63.84 $\pm$ 1.45} & 49.09 $\pm$ 0.95 \\
& 2D VoCo w/ extra unlabled data & \pmb{0.3817 $\pm$ 0.0019} & \pmb{61.91 $\pm$ 0.69} & \pmb{66.47 $\pm$ 1.09} & \pmb{54.88 $\pm$ 1.01} \\
\bottomrule
\end{tabular}
\end{table*}

\begin{table*}[tbh]
\centering
\caption{Classification results of EfficientNetV2-T on multi-organ and single-organ tasks}
\label{tab:multi_vs_single_v2t_by_task}
\begin{tabular}{cc|cccc}
\toprule
\textbf{Setting} & \textbf{Task} 
& \textbf{RSNA Score $\downarrow$} & \textbf{mAP (\%) $\uparrow$}
& \textbf{Precision (\%) $\uparrow$} & \textbf{Recall (\%) $\uparrow$} \\
\midrule
\multirow{2}{*}{w/o 2D VoCo}
& multi-organ   & 0.4133 $\pm$ 0.0068 & \pmb{59.47 $\pm$ 0.01} & \pmb{60.84 $\pm$ 0.55} & 49.37 $\pm$ 0.25 \\
& single-organ   & \pmb{0.3938 $\pm$ 0.0060} & 58.30 $\pm$ 0.70 & 54.28 $\pm$ 0.80 & \pmb{52.71 $\pm$ 1.90} \\
\midrule
\multirow{2}{*}{2D VoCo}
& multi-organ   & \pmb{0.3777 $\pm$ 0.0084} & \pmb{60.98 $\pm$ 1.09} & \pmb{66.71 $\pm$ 0.19} & \pmb{53.90 $\pm$ 0.15} \\
& single-organ   & 0.3966 $\pm$ 0.0060 & 58.69 $\pm$ 0.50 & 56.96 $\pm$ 1.50 & 53.80 $\pm$ 0.90 \\
\bottomrule
\end{tabular}
\end{table*}

\subsubsection{Effectiveness of 2D-VoCo Pre-training}
Our first experiment evaluated the impact of 2D-VoCo pre-training against the standard ImageNet-pre-trained baseline. As shown in Table~\ref{tab:voco_experiment2}, models pre-trained with 2D-VoCo (even when using only RATIC data) consistently outperformed the baseline (without 2D VoCo) on most metrics. This demonstrates that the features learned by our self-supervised approach, tailored to CT scan anatomy, provide a better initialization for the downstream task than standard ImageNet pre-training.

\subsubsection{Impact of External Unlabeled Data}
Table~\ref{tab:voco_experiment2} also clearly illustrates the benefit of leveraging additional unlabeled data. When pre-training was conducted on the combined RATIC + FLARE23 dataset (without using the FLARE23 labels), performance improved even further, establishing this as the best-performing setting. This confirms that pre-training on a larger, more diverse corpus of unlabeled data enables the model to learn more robust, generalizable features, leading to superior fine-tuning performance.

\subsubsection{Multi-Organ vs. Single-Organ Classification}
We then compared the performance of models trained on the entire abdominal scan (multi-organ) versus models trained on specific organ regions of interest (single-organ), as summarized in Table~\ref{tab:multi_vs_single_v2t_by_task}. For the baseline without the 2D VoCo model, results were mixed: the single-organ approach was slightly better on RSNA score and recall, while the multi-organ approach was better on mAP and precision.

However, a much clearer trend emerged for the 2D VoCo-pre-trained model (using the best RATIC + FLARE23 setting). Here, the multi-organ classification approach was definitively superior across all metrics. This strongly suggests that while the baseline model struggles to utilize the broad context of a full scan, the 2D-VoCo pre-training successfully teaches the model to leverage this contextual information and inter-organ relationships, rendering the multi-organ approach the clear and more efficient choice.

\section{Discussion}

In this study, we successfully adapted the 3D VoCo framework into a computationally efficient 2D version and demonstrated its effectiveness in enhancing multi-organ abdominal trauma classification. Our experimental results provide several insights into the value of self-supervised pre-training for complex medical imaging tasks.

The primary finding is that our 2D-VoCo pre-training consistently improves downstream classification performance over a standard ImageNet pre-trained baseline. The significant gains in various metrics suggest that the features learned by slice-level pre-training are more relevant to the anatomical context of CT scans than those learned from natural images.

Our comparison between multi-organ and single-organ classifications revealed the superiority of the former, which might be somewhat counterintuitive, as one might expect a model focused on a single organ's ROI to perform better. However, the superior performance of the multi-organ model suggests that it effectively leverages the broader context and inter-organ relationships present in the full abdominal scan. This finding has significant practical implications, as it eliminates the need for computationally expensive and potentially error-prone organ segmentation before classification, thus streamlining the entire automatic diagnostic pipeline.

Furthermore, the study highlights the benefit of including external unlabeled data during pre-training. The performance boost observed with the combined dataset indicates that the model learns more robust and generalizable features from a more diverse set of anatomical examples. This underscores the power of self-supervised learning to leverage large, readily available repositories of unlabeled medical data to overcome the limitations of smaller, annotated clinical datasets.

This research addresses the critical challenge of data scarcity in medical image analysis by proposing 2D-VoCo, a computationally efficient self-supervised learning framework adapted for slice-level pre-training. Our results demonstrate that this approach effectively learns rich, spatially-aware features from unlabeled CT scans, leading to significant performance improvements in the downstream task of multi-organ abdominal trauma classification. By successfully translating a complex 3D contrastive learning concept into a practical 2D workflow, this work presents a viable strategy for developing high-performing, memory-efficient models for clinical applications. Future work will explore more advanced sequence modeling architectures and improve model interpretability to further bridge the gap between research and clinical practice.

\section*{Acknowledgment and GenAI Usage Disclosure}

We acknowledge support from National Science and Technology Council of Taiwan under grant number 113-2221-E-008-100-MY3. 
The authors used ChatGPT and Gemini to improve language and readability. The authors reviewed and edited the content as needed and take full responsibility for the content of the publication.

\bibliographystyle{unsrt}  
\bibliography{ref}  

\end{document}